\documentclass{article}

\usepackage{xr}

\usepackage{listings}

\makeatletter
\newcommand*{\addFileDependency}[1]{% argument=file name and extension
\typeout{(#1)}% latexmk will find this if $recorder=0
\@addtofilelist{#1}
\IfFileExists{#1}{}{\typeout{No file #1.}}
}\makeatother

\usepackage{pgfplots}
\pgfplotsset{compat=1.15}
\usepgfplotslibrary{
  statistics,
  colorbrewer,
  groupplots,
}
\usetikzlibrary{
  patterns,
  shapes.geometric,
  decorations.text,
  matrix,
  fit,
  backgrounds,
  positioning,
}
\tikzset{
  fignode/.style={
    outer sep=0.25em,
  }
}
\tikzset{
  framedfignode/.style={
    outer sep=0.25em,
    inner sep=0.5em,
    rounded corners,
    draw,
  }
}
\colorlet{plotColorNeutral}{gray}
\definecolor{plotColor1}{HTML}{f61a1c}
\definecolor{plotColor2}{HTML}{377eb8}
\definecolor{plotColor3}{HTML}{4daf4a}
\definecolor{plotColor4}{HTML}{984ea3}
\colorlet{plotColorNeutral*}{plotColorNeutral!40}
\colorlet{plotColor1*}{plotColor1!60}
\colorlet{plotColor2*}{plotColor2!60}
\colorlet{plotColor3*}{plotColor3!60}
\colorlet{plotColor4*}{plotColor4!60}
\pgfplotsset{
    colormap={greenred}{HTML=(4daf4a) HTML=(e41a1c)},
    colormap={redgreen}{HTML=(e41a1c) HTML=(4daf4a)}
}
\usepackage[preprint,nonatbib]{neurips_data_2024}

\usepackage{graphicx}
\usepackage{subcaption}
\usepackage{tcolorbox}

\usepackage[utf8]{inputenc} % allow utf-8 input
\usepackage[T1]{fontenc}    % use 8-bit T1 fonts
\usepackage{hyperref}       % hyperlinks
\usepackage{url}            % simple URL typesetting
\usepackage{booktabs}       % professional-quality tables
\usepackage{amsfonts}       % blackboard math symbols
\usepackage{nicefrac}       % compact symbols for 1/2, etc.
\usepackage{tabularx}
\usepackage{longtable}
\usepackage{supertabular}
\usepackage{makecell}
\usepackage{colortbl}
\usepackage{multirow}       % simple URL typesetting

\definecolor{blue}{RGB}{17,220,247}
\definecolor{purple}{RGB}{163,115,250}
\definecolor{caribbeangreen}{rgb}{0.0, 0.8, 0.6}

\newcommand{\dexter}{\textsc{dexter}}

\newcommand{\COT}{\textsc{Few-shot-cot}}
\newcommand{\self}{\textsc{self-ask}}

\newcommand{\fshot}{\textsc{few-shot}}
\newcommand{\zfshot}{\textsc{zero-shot-cot}}
\definecolor{GREEN}{RGB}{84,130,53}
\newcommand{\colorit}{\cellcolor{green!15}}

\newcommand{\colorg}{\cellcolor{gray!15}}
\usepackage{microtype}      % microtypography
\usepackage{xcolor}         % colors

\title{DEXTER: A Benchmark for open-domain Complex Question Answering using LLMs}

\author{%
 Venktesh V$^*$, Deepali Prabhu$^*$, Avishek Anand  
\\    \texttt{{v.viswanathan-1,d.prabhu,avishek.anand}@tudelft.nl} 
\\Delft University of Technology\\
  Delft, Netherlands \\
}

\begin{document}

\maketitle
\def\thefootnote{*}\footnotetext{These authors contributed equally to this work}\def\thefootnote{\arabic{footnote}}
\begin{abstract}
Open-domain complex Question Answering (QA) is a difficult task with challenges in evidence retrieval and reasoning. The complexity of such questions could stem from questions being compositional, hybrid evidence, or ambiguity in questions. While retrieval performance for classical QA tasks is well explored, their capabilities for heterogeneous complex retrieval tasks, especially in an open-domain setting, and the impact on downstream QA performance are relatively unexplored. To address this, in this work, we propose a benchmark composing diverse complex QA tasks and provide a toolkit to evaluate state-of-the-art pre-trained dense and sparse retrieval models in an open-domain setting. We observe that late interaction models and surprisingly lexical models like BM25 perform well compared to other pre-trained dense retrieval models. In addition, since context-based reasoning is critical for solving complex QA tasks, we also evaluate the reasoning capabilities of LLMs and the impact of retrieval performance on their reasoning capabilities. Through experiments, we observe that much progress is to be made in retrieval for complex QA to improve downstream QA performance. Our software and related data can be accessed at \url{https://github.com/VenkteshV/DEXTER}.
\end{abstract}

\section{Introduction}

Addressing complex information needs expressed through complex questions is an active research area with many applications in critical areas such as finance and healthcare \cite{roy2022question:book,daull2023complex}. 
\begin{table*}[hbt!]
    \centering
    \small
    \begin{tabular}{p{2.5cm}
    p{2.7cm}cccccccp{0.3cm}p{0.3cm}p{0.3cm}}
    \toprule
     \textbf{Dataset}&{Complexity}& \multicolumn{1}{c}{Domain} &\multicolumn{1}{c}{Train} & \multicolumn{1}{c}{Dev} & \multicolumn{1}{c}{Test} & \multicolumn{1}{c}{Corpus Size} \\
   % &\multicolumn{1}{c|}{n@10} & \multicolumn{1}{c}{RR}&\multicolumn{1}{c|}{n@10} & \multicolumn{1}{c}{RR}&\multicolumn{1}{c|}{n@10} & \multicolumn{1}{c}{RR}&\multicolumn{1}{c|}{n@10} & \multicolumn{1}{c}{RR} &\multicolumn{1}{c|}{n@10} & \multicolumn{1}{c}{RR} &\multicolumn{1}{c|}{n@10} & \multicolumn{1}{c}{RR} &\multicolumn{1}{c|}{n@10} & \multicolumn{1}{c}{RR} \\
     % & cover-EM & cover-EM& EM& EM&EM \\
     \midrule
\midrule
    \colorg \textbf{Compositional} & \colorg &  \colorg& \colorg & \colorg& \colorg& \colorg \\
MusiqueQA (2-hop only) & Question& multi-domain&   14.3k&1.25k & 1.2k & 569,641 \\
     \colorg \textbf{Compositional + Comparative} & \colorg & \colorg & \colorg & \colorg& \colorg&  \colorg\\
WikiMultiHopQA & Question & multi-domain  &167.5k &12.6k &12.6k & 569,641\\
StrategyQA & Question&multi-domain  &2k&221 &490 &36.6M \\
     \colorg \textbf{Disambiguation} &\colorg  &\colorg  & \colorg &\colorg &\colorg &\colorg  \\
AmbigQA &Question,Answer &multi-domain  &10k &2k & 2k&24.3M 
 & \\

     \colorg \textbf{Table+Text} &\colorg  &\colorg  & \colorg &\colorg &\colorg  & \colorg \\
OTT-QA & Evidence&multi-domain  &41.4k &2.2k &2.1k &6.5M \\
  \colorg   \textbf{Table+Text}  \colorg & \colorg  &  \colorg &  \colorg & \colorg & \colorg & \colorg  \\
  \colorg  \textbf{+Numerical} \colorg & \colorg  &  \colorg &  \colorg & \colorg & \colorg & \colorg   \\
TAT-QA &Question, Evidence & Financial   &13.2k &1.6k &1.6k &7000 \\
FinQA & Question, Evidence & Financial &6.2k &883 &1147 &24.8k \\

     \bottomrule
    \end{tabular}
    \caption{Statistics of different datasets in \dexter{} benchmark.}
   % \vspace{-1em}
    \label{tab:statistics}
\end{table*}
The complexity usually stems from multiple aspects such as complexity due to compositional or ambiguous questions~\cite{2wikimultihopqa,musiqueqa,min-etal-2020-ambigqa,hotpotqa}, reasoning over heterogeneous evidence sources~\cite{tatqa,ottqa} and complexity due to answer formats~\cite{finqa,min-etal-2020-ambigqa}. 
A Question-Answering or QA system built to answer complex questions must have the capability to retrieve fine-grained information from multiple knowledge sources and perform reasoning over these gathered data to address the specific information needs. 
While several supervised and unsupervised approaches \cite{decomposing_1, zhou-etal-2022-learning-decompose, min-etal-2019-multi,decomposing_complex_multi_hop, khot2021textmodular, deng2022interpretable} have been proposed for the \textit{complex QA tasks}, a more recent paradigm of employing Large Language Models (LLMs) has emerged. 
Large Language Models are pre-trained on a large volume of data and are believed to encode world knowledge. They also exhibit abilities like In-Context Learning \cite{wei2022emergent,wei2023chainofthought,brown2020gpt3} performing tasks with few examples.

With recent advances made by  Large Language Models in several QA tasks \cite{self_ask,brown2020gpt3,trivedi-etal-2023-interleaving} they are being increasingly employed in a few-shot setting. 
This usually involves a closed book setting where the model is prompted to generate answers without access to external knowledge sources, by only relying on knowledge encoded in the model parameters. However, for complex QA tasks, a closed book setting is known to be challenging \cite{mdr,trivedi-etal-2023-interleaving} due to the need for multi-step reasoning, difficulty in question understanding and lack of required world knowledge. 
To circumvent this, Retrieval Augmented Generation (RAG) methods \cite{lewis2021retrievalaugmented} have been proposed, which augment the LLM with knowledge retrieved from external sources like Wikipedia or domain-specific collections. 
However, these RAG pipelines usually employ an off-the-shelf retriever \cite{min2024exploring,yu-etal-2023-unified,selfprompt_dpr} to mitigate training cost and generative models without evaluating the possible performance gaps in each of these components. 
For instance, an off-the-shelf dense retriever model that is pre-trained for passage retrieval might be sub-optimal for retrieving hybrid evidence like structured data encoded in tables.
Also, most pre-trained models are trained over a workload of simple questions, making them inadequate for compositional questions. 
Additionally, these systems have not been evaluated for \textit{complex QA} tasks, as prior works lack a comprehensive analysis of the retrieval performance of diverse models and the impact on downstream QA performance.

 In this work, we propose a heterogeneous benchmark \dexter{} covering diverse aspects of complexity such as complexity in questions, evidence sources, and answers as shown in Table \ref{tab:statistics}. Existing benchmarks either focus only on complex QA tasks concerning knowledge graphs\cite{wang2021benchmarking} or are limited to evaluating only answer generation or retrieval components specific to classical QA tasks \cite{wang2022modern,beir}. 
 Prior works in complexQA are also mostly focus only on limited aspects of complexity \cite{christmann2023compmix,islam2023financebench,wang2021benchmarking}. Additionally, individual benchmarks for complex QA assume access to gold standard evidence by evaluating the systems only in a reading comprehension setup \cite{musiqueqa,2wikimultihopqa,finqa,tatqa}. \dexter{} casts all datasets to an \textbf{open-domain setup} to evaluate retrieval in a more realistic setting. An open-domain setup is much more challenging as the retrieval model has to contend with distractors, and this problem is further exacerbated when dealing with hybrid evidence formats such as table and text.

 \dexter{} comprises \textbf{7 diverse datsets} covering different aspects of complexity and a corresponding multi-domain, large-scale collection to enable open-domain retrieval. \dexter{} offers a unified approach to benchmark retrieval and answer generation in an open-domain setting for complex QA tasks. 
 Usually, owing to the cost of training and lack of data, in many cases, pre-trained retrievers are directly employed. Hence, we evaluate \textbf{8} different pre-trained retrievers from lexical, sparse, dense, and late-interaction categories for context retrieval. Then, using the best retriever, we evaluate the ability of LLMs to generate answers. 
 We observe that there remains a large gap in retrieval for complex questions, providing scope for further research. Further, we observe that lexical methods like BM25 and late-interaction models like ColBERTv2 perform better than other dense retrieval models. Through extensive experiments, our goal is to answer the following research questions:

\textbf{RQ1}: How do off-the-shelf dense retrievers perform on the task of retrieving relevant context for open-domain QA tasks with different aspects of complexity?

\textbf{RQ2}: How well do LLM-based reasoning approaches perform for answering questions from different aspects of complexity in a closed-book setting?

\textbf{RQ3}: How well does augmenting LLM-based reasoning models with retrieved context perform on complex QA tasks in an open-domain setting?

\section{Related Work}
\label{sec:rel-work}
Complex Question Answering (CQA) is the task of answering questions requiring diverse abilities such as the ability to deal with hybrid evidence sources, numerical reasoning, common sense abilities, domain knowledge, and multi-step reasoning \cite{daull2023complex}. \dexter{} is an evaluation benchmark for complexQA tasks to evaluate retrieval and its impact on downstream QA performance. 
\vspace{-0.8em}
\subsection{Benchmarks}
To the best of our knowledge, \dexter{} is the first complexQA benchmark to evaluate pre-trained retrieval models in an \textit{open-domain setting} and downstream LLM-powered answer generation for diverse complex QA tasks. 
Prior benchmarks either focus on open-domain QA evaluation for classical QA tasks \cite{kamalloo-etal-2023-evaluating,wang2022modern,beir} or only concern with a narrow definition of complexity, such as reasoning over knowledge graphs \cite{wang2021benchmarking}.
BEIR\cite{beir} is a heterogeneous benchmark that only focuses on information retrieval performance for classical QA tasks and not its downstream impact on QA performance. The benchmark also does not cover diverse aspects of complexity. While they contain three datasets for Question Answering, namely NQ\cite{nq}, HotpotQA\cite{hotpotqa}, and FiQA-2018\cite{fiqa}, these datasets mostly contain corpus consisting of text passages and do not consider the complexity of retrieving hybrid evidence formats. 
Similarly, KILT\cite{kilt} is also an evaluation framework for knowledge intensive tasks which focuses on covering various tasks and domains and is not specific to Question Answering. Additionally, the QA tasks in KILT are primarily based on Wikipedia and do not consider the complexity of open-domain retrieval from sources with hybrid evidence such as table and text data. In contrast to the above two benchmarks, MultiReQA\cite{multireqa} is an evaluation benchmark specifically for Question Answering, however, they are limited to simple sentence-level retrieval tasks given a question. Additionally. this work has a limited focus on information retrieval, covering only two neural (BERT\cite{bert}, USE-QA\cite{useqa}) and one term-based retriever (BM25\cite{bm25}) and also only comprises \textit{small corpora} sizes with a maximum of 100K sentences. In our work, we aim to curate an evaluation benchmark that focuses on open-domain complex Question Answering covering diverse aspects of complexity over large corpus collections. %sOur benchmark curates 7 diverse datasets comprising various complexities. It incorporates a suite of 8 retrievers representing different classes of models.
% \subsection{Datasets}
% For our evaluation benchmark, we include 9 datasets comprising of diverse complexities - WikiMultiHopQA, StrategyQA\cite{strategyqa}, MusiqueQA\cite{musiqueqa}, OTT-QA\cite{ottqa}, TAT-QA\cite{tatqa} and FinQA\cite{finqa}. WikiMultiHopQA, StrategyQA and MusiqueQA contain questions that require multi-hop reasoning. Answering the questions in WikimultihopQA requires comparative multihop reasoning, StrategyQA needs implicit reasoning and Musique needs connected reasoning. Hence to fetch relevant evidence passages for these questions, the retriever requires various higher order abilities such as decomposition and iterative document fetching. AmbigQA contains ambiguous questions that can have multiple interpretations and hence multiple plausible answers. OTT-QA contains open domain questions that requires reasoning over both tables and text in a multi-hop fashion. Hence the retriever would need to perform cross modality open domain multi-hop retrieval. TAT-QA and FinQA also require information from both tables and text but are restricted to the financial domain. These questions necessitate numerical reasoning to derive the answer. Hence, the evidence retriever for such questions should be able to work with domain specific language and additionally deal with quantitative aspects. A summary of the above datasets and their properties are listed in Table.\ref{tab:main_result}
\begin{figure}
    \centering
    \includegraphics[width=0.751\textwidth]{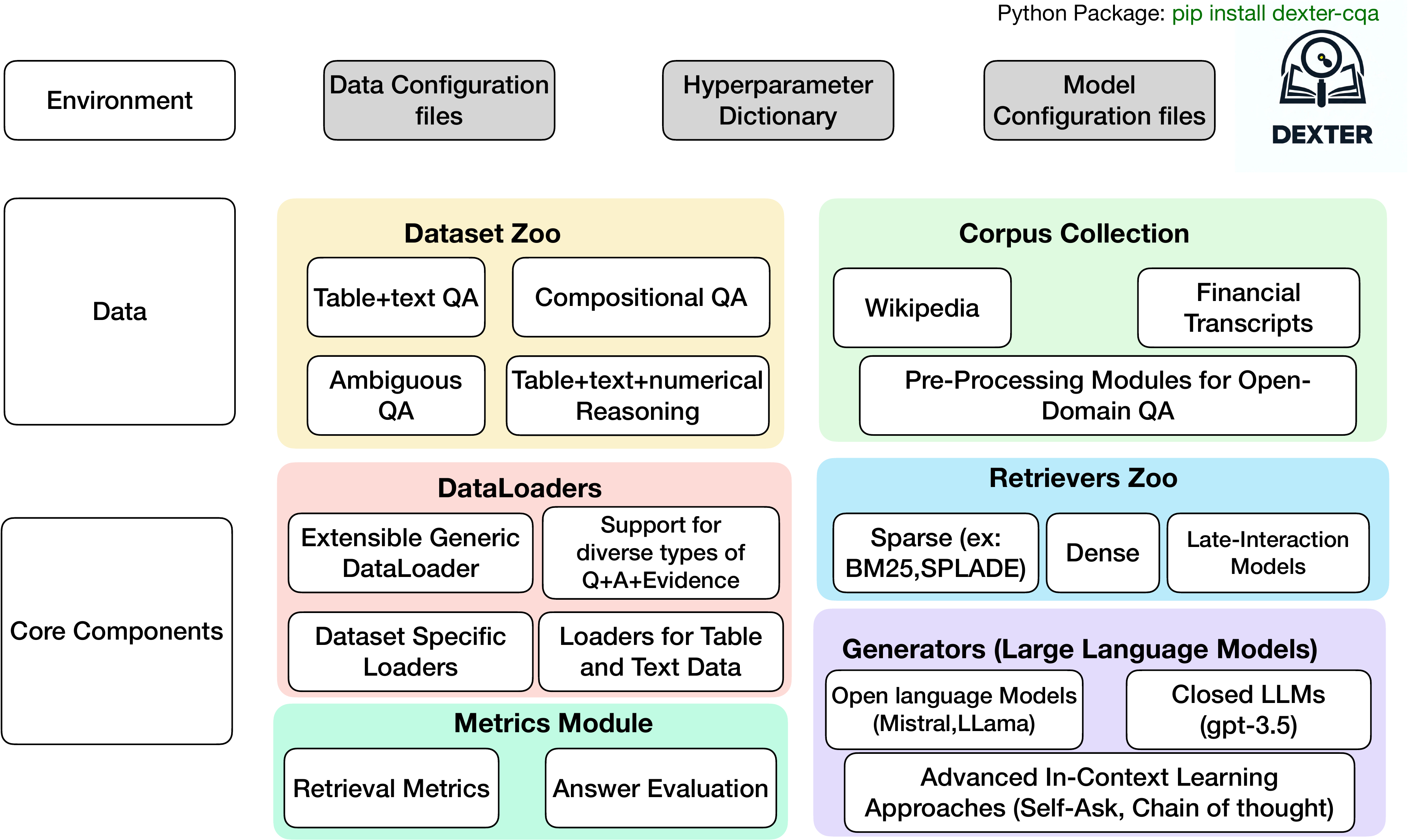}
    \caption{An Overview of \dexter{} Benchmark and ToolKit}
    \label{fig:enter-label}
\end{figure}

\subsection{Retrieval and Generative Components Evaluation for Complex Question Answering}

Question Answering usually entails retrieval of contexts from a knowledge source, followed by extractive or generative components to generate the final answer \cite{karpukhin-etal-2020-dense}. Existing solutions for complex QA primarily employ supervised and unsupervised approaches \cite{decomposing_1, zhou-etal-2022-learning-decompose, min-etal-2019-multi,decomposing_complex_multi_hop, khot2021textmodular, deng2022interpretable,multi_hop_dense_retrieval}. More recently, with the advances in Large Language Models \cite{brown2020gpt3,wei2022emergent,wei2023chainofthought}, a new paradigm of prompting LLMs to perform reasoning for addressing complex information needs has emerged. To steer the model's reasoning process, augmenting the LLM with external context has recently garnered interest with advances in retrieval augmented generation \cite{lewis2021retrievalaugmented}. However, many of the complex QA tasks do not have a realistic retrieval setup and tackle the task in a reading-comprehension setting \cite{musiqueqa,2wikimultihopqa,tatqa,finqa,ye2022unreliability}. Hence, in our work, we primarily cast all the datasets to an open-domain setting, which makes retrieval more challenging due to the presence of distractors. We also observe that existing works on some complex QA tasks usually employ an off-the-shelf retriever \cite{min2024exploring,yu-etal-2023-unified,selfprompt_dpr} to directly prompt generative models \cite{khot2023decomposed,self_ask,chen2022program} without evaluating the possible performance gaps in each of these components. Hence, existing works lack a comprehensive analysis of different retrieval methods across different complex QA tasks and their impact on downstream QA performance. 

\section{The \dexter{} Benchmark and Toolkit}
\label{sec:dexter-benchmark-toolkit}
The \dexter{} benchmark is designed to enable research in Complex Question Answering (CQA), encompassing different aspects of complexity. 
The toolkit accompanying the benchmark is designed in a modular manner to reduce the entry barrier for researchers while also enabling extension to new complex QA tasks and methods. 
Figure \ref{fig:enter-label} depicts the core modules and features in the \dexter{} toolkit for benchmarking retrieval and answer generation components. One of our main contributions is also evaluating the retrieval models for complex QA tasks in an \textbf{open-domain} setting. More details on the open-domain setting is provided in Section \ref{label:datasets}. \dexter{} offers a unified benchmark to evaluate pre-trained retrieval models and answer generation models in a principled manner.

\subsection{Datasets}
\label{label:datasets}
The prior benchmarks, primarily focus on limited aspects of complexity such as reasoning over knowledge graphs or multi-hop QA \cite{christmann2023compmix,lan2021survey,multi_hop_survey}.  However, \dexter{} covers diverse dimensions of complexity such as (i) Complexity due to \textit{Compositional and Comparative} reasoning (MusiqueQA \cite{musiqueqa}, 2WikiMultiHopQA \cite{2wikimultihopqa}), (ii) Complexity due to \textit{implicit and common-sense reasoning} based questions (StrategyQA \cite{strategy_qa}) (iii) Complexity due to \textit{ambiguity} in questions (AmbigQA \cite{min-etal-2020-ambigqa}) and (iv) complexity due to reasoning over \textit{different modalities / heterogeneous sources} such as table and text (TAT-QA \cite{tatqa}, OTT-QA \cite{ottqa}) and (v) complexity due to \textit{numerical reasoning} over heterogeneous sources (FinQA \cite{finqa}). 

MusiqueQA and 2WikiMultihopQA require multihop-reasoning for retrieval around different aspects of the question, followed by reasoning over the retrieved contexts. StrategyQA requires deciphering implicit aspects in the question for retrieval and commonsense reasoning around these aspects. AmbigQA contains ambiguous questions that can have multiple interpretations and hence multiple plausible answers. TAT-QA, FinQA, and OTT-QA require retrieval over heterogeneous evidence sources such as tables and text followed by compositional or numerical reasoning. 

Hence, it is evident that the discussed aspects of complexity require retrieval from knowledge sources followed by reasoning over the contexts. However, we observe that the majority of the baselines to these datasets are designed in a reading comprehension setup \cite{finqa,chen2022program,musiqueqa,tatqa} which assumes access to gold contexts for a given question or relies on prompting generative models for knowledge \cite{self_ask,khot2023decomposed,ye2022unreliability} without comprehensively evaluating retrieval. However, in the real world, the QA pipelines for complex QA would have to contend with retrieval from large corpus collections with \textit{distractors}. For a more realistic evaluation, we cast the retrieval task for these datasets to an open-domain setup. 

\textbf{Open Domain Setup}: The compositional and comparative reasoning datasets 2WikimultihopQA (WQA) and MusiqueQA (MQA) have a reading comprehension setup where each question is paired with 10 to 20 paragraphs. To convert them to an open-domain setup, we combine the paragraphs from all questions. This results in 430,225 paragraphs for 2WikiMultihopQA and 139,416 for MusiqueQA. We combine these together to enable a realistic open-domain setting with distractors resulting in 569,641 paragraphs. For query relevance judgments also known as \textit{qrels} which indicate relevant contexts for a question to evaluate retrieval, we use the contexts given for each question.

For strategyQA we use the Wikipedia dump with 36.6 million passages used to collect the dataset. For \textit{qrels}, we use the title information of Wikipedia passages that annotators looked up when annotating supporting facts for each question and use this to extract corresponding passages from the Wikipedia dump. The matching passages are graded as 1 indicating their relevance to the query. The resulting qrels are used to evaluate the retrieval setup.

While AmbigQA was also constructed using Wikipedia, the original dataset does not come with annotated passages for each question. We use the Wikipedia dump used for annotation as a corpus comprising 24.3M passages. For constructing qrels we use the semi-oracle information provided by authors of AmbigQA \cite{min-etal-2020-ambigqa}. This information points to the Wikipedia articles clicked and viewed by annotators when annotating answers to a question. Due to multiple interpretations of the question, this list comprises multiple passages. We use those passages that were viewed by the annotators and contain valid answers to the question as positive and relevant passages to the questions in the qrels. The rest are considered as negative passages as they do not contain a valid answer, though may be lexically related to the question.

For FinQA, we use the financial transcripts used to generate the questions as the evidence collection. FinQA was originally intended to be a reading comprehension dataset paired with a table and a text passage for each question. We leverage this to form the qrels for retrieval evaluation. We use a similar process for TAT-QA.
The original work of OTT-QA had already released an open-domain set up with a corpus collection of Wikipedia passages paired with corresponding Wikipedia tables. While the original work \cite{ottqa} does not perform an extensive evaluation of the retrieval step, \dexter{} evaluates the ability of diverse models to retrieve from heterogeneous sources with different modalities of evidence. For all datasets with hybrid evidence, we serialize the table by separating \textbf{columns} using the "|" symbol and \textbf{rows} are separated by a newline to enable encoding by transformer-based models.

\vspace{-0.9em}
\subsection{\dexter{} Software Framework}
\vspace{-0.5em}
\label{sec:software}
The \dexter{} software\footnote{\url{https://github.com/VenkteshV/DEXTER}} is available as a python package (pip install dexter-cqa) to enable ease of use. Our setup doubles as a toolkit to benchmark retrieval models and LLM-based reasoning for complex QA tasks. We provide customizable data loaders for different data modalities (table+text), typed model classes, pre-processing scripts for corpus formation, and wrappers for customizable retrieval models and LLMs for inference as shown in Figure \ref{fig:enter-label}. We also provide an implementation of standard metrics for evaluating retrieval and answer generation. We adhere to SOLID principles of software development to enable the extension of the benchmark to new datasets, retrieval, and generative models. \dexter{} implements a generic data loader that is customizable by class extension. Our environment is driven by configuration files where the user can specify the dataset, corpus, and model names to use for an experiment. The \textit{Orchestrator} module invokes the corresponding classes to create instances and helps in running retrieval and answer generation in a sequential manner. All our data is stored in JSON format. We also support the caching of retrieval results in JSON format. This enables using retrieval outputs from non-open-source retrieval models, where the user can input the JSON of retrieval results as input to the answer generation model. Overall, \dexter{} is a modular, unified toolkit enabling benchmarking of pre-trained retrieval models and answer generation through generative models for complex QA tasks.
%\vspace{-1em}
\subsection{Evaluation Metrics}
We implement the Metrics module as shown in Figure \ref{fig:enter-label}.

\textbf{Retrieval}: While \dexter{} provides several evaluation metrics for retrieval, we choose Normalized Discounted Cumulative Gain (nDCG@k) as the primary metric for our results. This is because nDCG is a rank-aware metric that is well-suited for binary and graded relevance judgments. Precision and recall metrics are rank-unaware and hence not well suited to comparison across diverse complex QA tasks. It also has nice theoretical advantages, as discussed in the work \cite{wang2013theoretical}. In our benchmark, we provide a wrapper over the Python interface of the TREC eval tool \cite{VanGysel2018pytreceval}.

\textbf{Answer generation}: For evaluating the generative models used for answer generation in \dexter{}, we employ several metrics based on the complex QA task. We use \textit{Cover-EM (c-EM) }\cite{self_ask,rosset2021knowledgeaware} for MusiqueQA, 2WikiMultiHopQA, TAT-QA, OTT-QA and StrategyQA following the prior works. Cover-EM checks if the gold answer is a substring of the generated answer. Since LLMs may generate additional text such as explanation/reasoning in addition to the exact answer, this metric is more suited for complex QA tasks that comprise free-form answers of different types. For instance, for a question \textit{``What position was held by Warren Hastings?"} while a ground truth answer could be \textit{``Governor-general"}, a generated answer could be \textit{``Governor-general of Bengal"}. For FinQA, we follow the work POT \cite{chen2023program} and implement a metric \textit{EM-tol} which performs a match between predicted and ground truth answers by using \textit{math.isclose} with a relative tolerance of 0.02. This is a relaxation of strict Exact Match for numerical answers, as LLMs cannot generate high-precision floats and large numbers, which is a characteristic of FinQA.
\vspace{-0.6em}

\section{Experimental Setup}
\label{sec:experiments}
We use \dexter{} to benchmark diverse state-of-the-art retrieval approaches for context retrieval in an open-domain setup and state-of-the-art generative LLMs for answer generation. Our experiments are carried out on a server with two GeForce RTX 3090 GPUs with a combined GPU memory of 48 GB.
\vspace{-0.6em}
\subsection{Retrievers}
 We focus primarily on transformer based architectures and leverage publicly available checkpoints as shown in Table \ref{tab:model_char} in Appendix \ref{appendix:metadata}. We evaluate retrieval models from several categories. However, \dexter{} is model agnostic, and new models can be easily added because of the modular framework. For all datasets except FinQA we evaluate on dev set due to lack of publicly available test sets with gold labels. For FinQA we evaluate on test set.
 
\textbf{Lexical retrieval}: BM25 \cite{bm25} computes token-matching between sparse representations using TF-IDF weights. We employ ElasticSearch's BM25 implementation due to its enhanced query speed and maintainability over other implementations\cite{fogelberg2023search}.

\textbf{Sparse retrieval}: For sparse retrieval we employ SPLADE \cite{splade,SPLADEv2} which is a neural model employing query and document sparse expansions. %SPLADE accomplishes this by using the BERT Masked Language Modeling head and additionally employing sparse regularization. 

\textbf{Dense retrieval}: DPR\cite{karpukhin-etal-2020-dense} trains a dual encoder model with large datasets to maximize similarity between related queries and documents. We found the open source model trained on multiple datasets \textit{facebook-dpr-question-encoder-multiset-base} to work better than the model trained on individual datasets like the bi-encoder trained on Natural Questions. We also include ANCE\cite{ance} in our evaluation which is a bi-encoder model that samples hard negatives through an Approximate Nearest Neighbour search over an index of the corpus which is also updated yielding better negatives. We employ the checkpoint \textit{msmarco-roberta-base-ance-firstp} trained on MS-MARCO \cite{bajaj2018ms}. Tas-b \cite{tas-b} is a bi-encoder model trained using supervision from a cross-encoder and ColBERT.
%takes it once step further by incorporating global negatives during training to improve accuracy and convergence speed. BERT which serves as the encoder for both ANCE\cite{ance} and DPR\cite{karpukhin-etal-2020-dense}, can neglect dependencies between predicted tokens. To overcome this, MPNet\cite{mpnet} uses a combination of Masked Language Modelling and Permutated Language modelling to enhance contextual understanding.

\textbf{Late-Interaction Models}: We implement the late-interaction model ColBERTv2 \cite{colbert,santhanam-etal-2022-colbertv2} in \dexter{}. It is a multi-vector late-interaction model that employs a cross-attention-based MaxSim operation to capture fine-grained interactions between query and document token representations.
\begin{table*}[hbt!]
    \centering
    \small
    \begin{tabular}{lccccccccccccccc}
    \toprule
     \textbf{Method}&\multicolumn{1}{c}{WQA} & \multicolumn{1}{c}{MQA} & \multicolumn{1}{c}{AmbigQA} & \multicolumn{1}{c}{StrategyQA}& \multicolumn{1}{c}{Tat-QA} &\multicolumn{1}{c}{ FinQA} &\multicolumn{1}{c}{ OTT-QA} \\
   % &\multicolumn{1}{c|}{n@10} & \multicolumn{1}{c}{RR}&\multicolumn{1}{c|}{n@10} & \multicolumn{1}{c}{RR}&\multicolumn{1}{c|}{n@10} & \multicolumn{1}{c}{RR}&\multicolumn{1}{c|}{n@10} & \multicolumn{1}{c}{RR} &\multicolumn{1}{c|}{n@10} & \multicolumn{1}{c}{RR} &\multicolumn{1}{c|}{n@10} & \multicolumn{1}{c}{RR} &\multicolumn{1}{c|}{n@10} & \multicolumn{1}{c}{RR} \\
     % & cover-EM & cover-EM& EM& EM&EM \\
     \midrule

    \midrule
    
    \textbf{Lexical} & & \\
            BM25 \cite{bm25} &\textbf{0.327} & 0.191& \textbf{0.316}&0.101&0.432 & 0.155 & 0.149 \\

     \midrule
     
      \textbf{Sparse} & & \\
            SPLADEV2 \cite{SPLADEv2} & 0.251&0.155 & 0.268 & 0.087& 0.355& 0.118& 0.107 \\
     \midrule
      \textbf{Dense} & & \\
            DPR \cite{karpukhin-etal-2020-dense} & 0.126& 0.109&0.135&0.042  &0.212&0.052&0.058      \\
            ANCE \cite{ance} & 0.212&0.140 & 0.272 & 0.091&0.287&0.086&0.062 \\

                  %      MDR & & \\
                        tas-b \cite{tas-b} & 0.277& 0.176&0.275&0.126&0.349&0.099&0.096\\
                        MPNet \cite{mpnet} & 0.222 & 0.163 & 0.193& 0.127 & 0.323 & 0.103&0.129\\
       Contriever \cite{contriever} & 0.216 & 0.155&0.149&0.053&0.164&0.059&0.062 \\
    COlBERTV2 \cite{santhanam-etal-2022-colbertv2} & 0.294 &\textbf{0.199}&0.297&\textbf{0.127}&\textbf{0.433}&\textbf{0.155} & \textbf{0.196} \\
\midrule
% \textbf{LLM+Retrieval} \\
%         Decompose-retrieve & &&&&& \\
%     \midrule

    % \textbf{Re-Ranking} & & & & & & & \\
    % BM25 + CE & & & & & & & \\
    % \bottomrule
%          \textbf{Semi-Oracle} & & & & & & & & & & \\
%         ClaimOnly & 33.03& 39.57& 36.31 & 58.15 & 33.81& 48.61& 25.70 & 23.99&28.55 & 63.79 & 7.95 & 33.42 &  43.70  \\

% \programfc{} & 38.57& 42.49 & 37.12  & 50.66& 35.22& 45.76& 33.43&32.50 & 32.95 & 55.11 & 25.44 & 37.83 & 43.79 \\
%     \claimdecomp{}&33.43 & 39.78 & 35.04&  55.49 & 33.93 & 48.53& 34.37& 33.50&29.48 & 63.11& 10.85 & 34.48& 44.19  \\
%      \numdecomp{}& 33.81& 39.46 &33.57 & 53.45& 34.18& 47.00& 35.23&34.23 &29.11 & 60.29 & 13.90 &34.43& 43.24\\
    \end{tabular}
    \caption{Retrieval results on \dexter{}, nDCG@10 across datasets.}
     \vspace{-1em}
    \label{tab:main_result}
\end{table*}

\subsection{Generative Models}
We support various open generative models such as LLama2, Mistral, and FLANT5-XL. We also support OpenAI models. In our experiments, we primarily employ gpt-3.5-turbo among the OpenAI models. For our main experiments, during LLM inference, we set the maximum number of output tokens to 512, and temperature to $0.3$ to mitigate randomness, with frequency and presence penalty set to 0.8 and $0.6$ to avoid repetition. The prompts employed can be found in Appendix \ref{appendix:prompts}.

\section{Results}
\label{sec:results}
The retrieval results across complex QA datasets are shown in Table \ref{tab:main_result}. We also report Recall@100 in Appendix \ref{appendix:recall}, Table \ref{tab:recall_100}. The answer generation results are shown in Table \ref{tab:llm_reasoning}.
%From the results, we gather that Large Language models are yet to excel at reasoning over hybrid evidence sources and modelling ambiguity. Additionally, off-the-shelf performance of diverse dense retrievers on retrieving context for  diverse complex QA tasks fall short, affecting downstream QA performance.
 \vspace{-0.7em}
\subsection{Comparison of Retrieval Models on Diverse Complex QA Tasks}
To Answer \textbf{RQ1}, we evaluate lexical, late-interaction, and diverse dense-retrieval models on the curated datasets in an open-domain setup. From Table \ref{tab:main_result}, we observe that the lexical model BM25 is a strong baseline across all complex QA tasks, including reasoning over hybrid sources. We observe that DPR has the lowest performance, indicating that while this model is suited for simpler open-domain QA tasks, it falls short on complex QA tasks. We observe that though DPR was trained on Natural Questions it falls short on AmbigQA which is a derivative of natural questions. We posit that this is because the bi-encoder model was not trained to handle ambiguity, indicating that complex QA requires further care compared to traditional questions.  They fall short on tasks that require multistep reasoning during retrieval or when retrieving hybrid evidence like tables and text from multiple sources. Surprisingly, we also observe that models like \textit{contriever} that are pre-trained for retrieval underperform compared to lexical models like BM25 and other dense retrieval models like tas-b and MPNet. We also observe that neural sparse models like SPLADE are competitive or even outperform dense retrieval models and prove to be strong baselines following BM25.

Late-Interaction model ColBERT (v2) \cite{santhanam-etal-2022-colbertv2} outperforms other methods on 5/7 datasets. We posit that cross-attention-based operation on tokenwise representations helps capture intricate relationships between query and context and also between different modalities of evidence such as table and text. This aids in higher retrieval performance on datasets like StrategyQA that require implicit reasoning and also on all table and text-based QA datasets compared to other dense retrieval models. We also find similar observations in individual works like \cite{zhong2022reasoning} which employs cross-attention fusion between table and text evidence and \cite{multi_hop_dense_retrieval} which employs cross-attention for multi-hop reasoning.
\begin{table*}[hbt!]
    \centering
    \small
 \begin{tabular}{lccccccccccc}
    \toprule
     \textbf{Method}& \multicolumn{1}{c}{MQA}& \multicolumn{1}{c}{WQA} & \multicolumn{1}{c}{AmbigQA} &\multicolumn{1}{c}{StrategyQA} &\multicolumn{1}{c}{Tat-QA} &\multicolumn{1}{c}{FinQA} & \multicolumn{1}{c}{OTT-QA}\\

      & c-EM & c-EM& $F1_{Ans}$&c-EM&c-EM & EM-Tol & c-EM \\
     \midrule
     \colorg \textbf{Closed-book QA} & \colorg & \colorg & \colorg &\colorg  &\colorg  & \colorg & \colorg   \\
     \fshot{} \cite{brown2020gpt3}& 11.66& 28.41& 14.28&62.88 & -& - &-\\
     
     \COT{} \cite{wei2023chainofthought} &20.85& 34.08& 14.55& 66.38 & - &-&- \\
     
     \zfshot{} \cite{kojima2023large} & 12.22& 19.25& 13.00 & 25.33& -& - & - \\
    \hline
    \self{} \cite{self_ask} & 23.64& 32.58& -& 68.99& - &-  & - \\
        \midrule 
     
\colorg \textbf{Open-domain QA} & \colorg & \colorg & \colorg &\colorg  &\colorg  & \colorg & \colorg   \\
 
    \textbf{RAG-Oracle} & & & & & & & \\ 
         \fshot{}& 41.53 & 58.17 &32.24 & 72.93 & 50.75 & 42.55 & 40.53\\
\COT{} & \\
        \ \ - gpt-3.5-turbo & \colorit \textbf{44.28} & \colorit \bf 65.55& 35.57  &  \colorit \bf73.36& 54.05&52.22 & \colorit \bf 46.23 \\ 
                  \ \ - Mistral-7b & 22.54& 25.00 &09.93 & 62.55 &42.99 &12.55 &  18.39 \\ 
%           \ - Llama2-7b  &  & & & & &06.62 & \\
   \zfshot{} & 33.87 & 60.75& 27.09 &35.37 & 34.25& 47.51& 42.21 \\ 
 
   \midrule
       \textbf{RAG} & & & & & & & \\ 
                \fshot{}& 20.93& 38.58 & 20.08& 62.01& 24.04 & 16.83 &8.72 \\

       \COT{} & 25.80 & 47.33 & 27.37 & 63.32 & 25.78 & 17.44 & 10.61 \\ 
   \zfshot{} & 20.93 & 32.58 &12.60&  51.09 & 20.44 & 16.48 & 10.07 \\
\midrule
    supervised baseline &  37.60  & 50.59 &  \colorit \textbf{42.30}  & 65.2   & \colorit \bf55.2 & \bf \colorit  58.86 & 33.20\\
     
     \bottomrule
    \end{tabular}
    \caption{QA Performance on \dexter{}. The model used  is gpt-3.5-turbo, unless otherwise specified.}
    % \vspace{-1cm}
    %$\dagger$ and $\ddagger$ indicates statistical significance over \COT{} at 0.1 level  and 0.01 level respectively.
    \label{tab:llm_reasoning}
    \vspace{-2em}
\end{table*}

\textbf{Insight 1}: \textit{Pre-trained dense retrieval models fall short on retrieval performance for compositional questions and when retrieving from hybrid knowledge sources. In contrast, lexical models serve as strong baselines and late-interaction models demonstrate significant potential. }
\vspace{-0.7em}
\subsection{LLM Reasoning Performance on Complex QA Tasks in Closed Book Setting}
 To answer \textbf{RQ2}, we evaluate the performance of Large Language Models on diverse complex QA tasks in a closed-book setting where the models are expected to answer questions without access to the context. Since TAT-QA, OTT-QA, and FinQA have questions that are tightly coupled to context, they are incomprehensible and unanswerable without context and hence cannot be evaluated in a closed-book setting. We evaluate various state-of-the-art prompting strategies such as chain-of-thought \cite{wei2023chainofthought} and self-ask \cite{self_ask} which have shown improvements for complex reasoning tasks. We observe that though models like gpt-3.5-turbo are believed to encode world knowledge, they overall underperform on compositional reasoning tasks when not provided with relevant context. While chain-of-thought and self-ask improve compositional and comparative reasoning performance over few-shot prompting by eliciting reasoning chains from LLMs, they are still limited without access to context. However, we observe that in StrategyQA which requires commonsense reasoning and reasoning over implicit aspects, gpt-3.5-turbo achieves impressive gains. This can be ascribed to the factual knowledge encoded in the model parameters due to pre-training on Wikipedia passages. These passages serve as a common source for annotators to construct supporting facts answering the questions. We also observe that emergent capabilities like chain-of-thought are more pronounced in large-scale models like gpt-3.5-turbo leading to significant performance improvements than smaller models like Mistral-7b (In Table \ref{tab:llm_reasoning}) or LLama2-7b (In Appendix \ref{appendix:open_llm}). 
 
\textbf{Insight 2}: \textit{The retrieval plays an important role in complex QA tasks. LLMs do not encode sufficient parametric knowledge to solve complex QA tasks in a closed-book setting.}
\vspace{-0.8em}
 \subsection{LLM Reasoning Performance in Retrieval Augmented Open-Domain Setting}
 To answer \textbf{RQ3}, we carry out experiments in two different setups. To obtain an upper bound on LLM reasoning performance, we carry out an answer generation task in an Oracle setup. In an Oracle setup, the LLMs are fed with gold contexts as input. We observe that in compositional and comparative reasoning tasks, the performance of the model increases significantly compared to the closed-book setting and even supervised state-of-the-art methods. We observe that for commonsense reasoning, providing relevant context to LLM in \zfshot{} prompting setting further improves results compared to the closed-book setting. This reinforces the need for augmenting relevant context for addressing complex information needs, even for LLMs that are believed to encode world knowledge in model parameters.

 We observe that for reasoning hybrid contexts such as table and text data, Large Language Models still fall short compared to state-of-the-art supervised approaches even when provided with gold contexts in the Oracle setting. While some existing results have shown the capabilities of LLMs to be tabular reasoners \cite{tableQALLM}, our results highlight the gaps and need for further enhancements to support hybrid data modalities in large generative models.

  \textbf{Insight 3}: \textit{Retrieving relevant context for complex QA tasks has significant performance gaps. Large Language Models have significant performance gaps when reasoning over hybrid evidence sources and cannot sufficiently model ambiguity in questions even in the presence of gold contexts.}
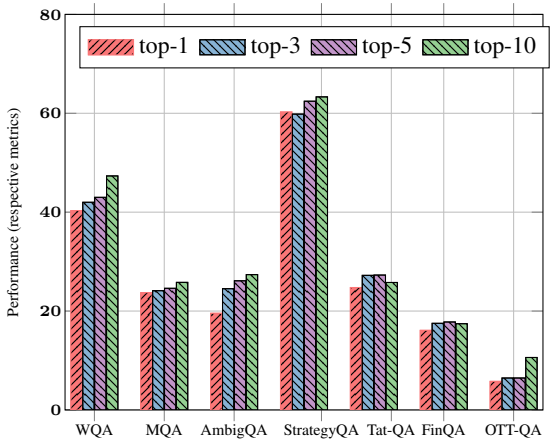
\begin{figure}[hbt!]
    \centering
    \begin{subfigure}{.71\linewidth}
    \centering
    \begin{tikzpicture}
\edef\mylst{"69.37","48.82","76.78","54.75"}
\edef\explora{"79.07","53.94","79.65","59.46"}

    \begin{axis}[
            ybar=0.2pt,
            width=0.8\textwidth,
            bar width=0.65,
            every axis plot/.append style={fill},
            grid=major,
            xtick={1, 5, 9, 14,18,21,25},
            xticklabels={WQA, MQA, AmbigQA, StrategyQA,Tat-QA,FinQA,OTT-QA},
            xlabel={},
            ylabel style = {font=\tiny},
        yticklabel style = {font=\boldmath \tiny,xshift=0.5ex},
        xticklabel style ={font=\tiny,yshift=0.5ex},
            ylabel={Performance (respective metrics)},
            enlarge x limits=0.07,
            ymin=0,
            ymax=80,
            legend style ={font=\small,yshift=0.01ex},
            area legend,
            nodes near coords style={font=\tiny,align=center,text width=1em},
            legend entries={top-1, top-3, top-5, top-10},
            legend cell align={left},
            legend pos=north west,
            legend columns=-1,
            legend style={/tikz/every even column/.append style={column sep=0.06cm}},
        ]
        \addplot+[
            ybar,
            plotColor1*,
            postaction={
                    pattern=north east lines
                },
        ] plot coordinates {
                (1,40.33)
                (5,23.72)
                (9,19.55)
                (13,60.26)
                (17,24.76)
                (21,16.10)
                (25,5.78)

            };
        \addplot+[
            ybar,
            plotColor2*,
            draw=black,
    nodes near coords align={vertical},
            postaction={
                    pattern=north west lines
                },
        ] plot coordinates {
                (1,42.00)
                (5,24.12)
                (9,24.52)
                (13,59.83)
                (17,27.22)
                (21,17.52)
                (25,6.46)
            };
         \addplot+[
            ybar,
            plotColor4*,
            draw=black,
    nodes near coords align={vertical},
            postaction={
                    pattern=north west lines
                },
        ] plot coordinates {
                (1,43.00)
                (5,24.60)
                (9,26.12)
                (13,62.44)
                (17,27.28)
                (21,17.79)
                (25,6.46)
            };
         \addplot+[
            ybar,
            plotColor3*,
            draw=black,
    nodes near coords align={vertical},
            postaction={
                    pattern=north west lines
                },
        ] plot coordinates {
                (1,47.33)
                (5,25.80)
                (9,27.37)
                (13,63.32)
                (17,25.78)
                (21,17.44)
                (25,10.61)
            };
    \end{axis}
\end{tikzpicture}
    \end{subfigure}
    \caption{Effect of \# of retrieved docs. using ColBERTv2 on QA perf. (few-shot-cot,gpt-3.5-turbo)}
    \label{fig:plot}
    \end{figure}

\vspace{-1.4em}
\subsection{Impact of Number of Retrieved Contexts on Answer Generation performance}

To analyze the effect of the number of retrieved contexts on downstream QA performance, we perform experiments using top-1, top-3, top-5, and top-10 retrieved contexts (Figure \ref{fig:plot}). We employ ColBERTv2 as the retrieval model and gpt-3.5-turbo as the generative model. The plot shows the perils of choosing an insufficient number of documents. For instance, on WQA with top-1 context, the performance drops by 17.36\% compared to reasoning with top-10 contexts, and on OTT-QA performance drops from 10.61 to 5.68 (by about 83\%). Usually, retrieving more contexts increases the chance of retrieving distractors, which might impact performance. However, since complex QA tasks require multi-step reasoning over multiple contexts, we posit that retrieving more contexts helps the LLM to reason around different aspects of the question with respect to diverse contexts. Additionally, it has been observed that distractors in retrieved contexts also aid the LLM in contrastive reasoning by contrasting information from relevant and distractor contexts \cite{zhao2024enhancing}. We also observe that retrieving excessive contexts also deteriorates performance, as observed on TAT-QA where performance drops by 5.81\% for top-10 contexts.  We observe that this is primarily because TAT-QA and FinQA requires extracting precise numerical information from financial documents, and distractors would result in incorrect reasoning.
\vspace{-0.85em}
\section{Conclusion}
In this work, we presented \dexter{}, a benchmark for complex QA with a  coverage of diverse aspects of complexity and data in easy-to-use formats with a modular toolkit to evaluate retrieval and generative models over hybrid evidence sources in an \textit{open-domain} setting. We observe that lexical models like BM25 and sparse retrievers serve as strong baselines. However, overall we observe that retrieval has immense scope for improvement in the presence of ambiguous or compositional questions and when retrieving hybrid evidence such as table and text. We also observe that while Large Language models are believed to encode world knowledge, they grossly underperform on complex QA tasks in a closed-book setting. Large Language Models also lack at modeling ambiguity and demonstrate gaps in parsing information from hybrid evidence formats such as table and text, even when provided with gold evidence. We hope \dexter{} will help further research in complex Question Answering to address these gaps.

\bibliography{ref}
\bibliographystyle{plain}

%%%%%%%%%%%%%%%%%%%%%%%%%%%%%%%%%%%%%%%%%%%%%%%%%%%%%%%%%%%%
\section*{Checklist}

%%% BEGIN INSTRUCTIONS %%%
% The checklist follows the references.  Please
% read the checklist guidelines carefully for information on how to answer these
% questions.  For each question, change the default \answerTODO{} to \answerYes{},
% \answerNo{}, or \answerNA{}.  You are strongly encouraged to include a {\bf
% justification to your answer}, either by referencing the appropriate section of
% your paper or providing a brief inline description.  For example:
% \begin{itemize}
%   \item Did you include the license to the code and datasets? \answerYes{See Section~\ref{gen_inst}.}
%   \item Did you include the license to the code and datasets? \answerNo{The code and the data are proprietary.}
%   \item Did you include the license to the code and datasets? \answerYes{}. See Appendix \ref{sec:license} for more details 
% \end{itemize}
% Please do not modify the questions and only use the provided macros for your
% answers.  Note that the Checklist section does not count towards the page
% limit.  In your paper, please delete this instructions block and only keep the
% Checklist section heading above along with the questions/answers below.
%%% END INSTRUCTIONS %%%

\begin{enumerate}

\item For all authors...
\begin{enumerate}
  \item Do the main claims made in the abstract and introduction accurately reflect the paper's contributions and scope?
    \answerYes{}. Please see Abstract and Introduction.
  \item Did you describe the limitations of your work?
    \answerYes{} We discuss limitations of \dexter{} benchmark in Appendix \ref{appendix:limitations}.
  \item Did you discuss any potential negative societal impacts of your work? \answerNo{} We focus on benchmarking retrieval and QA performance for complex QA tasks. 
 \item Have you read the ethics review guidelines and ensured that your paper conforms to them?
     \answerYes{}

\end{enumerate}

\item If you are including theoretical results...
\begin{enumerate}
  \item Did you state the full set of assumptions of all theoretical results?
    \answerNA{}
	\item Did you include complete proofs of all theoretical results?
    \answerNA{}
\end{enumerate}

\item If you ran experiments (e.g. for benchmarks)...
\begin{enumerate}
  \item Did you include the code, data, and instructions needed to reproduce the main experimental results (either in the supplemental material or as a URL)?
    \answerYes{}. URL is mentioned in Abstract and Section \ref{sec:software}. Further details are provided in Section \ref{appendix:metadata}.
  \item Did you specify all the training details (e.g., data splits, hyperparameters, how they were chosen)?
    \answerYes{} All results are reproducible using the source code in the GitHub repository mentioned in the abstract. We also explicitly mention hyperparameters and inference settings in Section \ref{sec:experiments}, in Appendix \ref{appendix:prompts}, Appendix \ref{appendix:experimental}.
	\item Did you report error bars (e.g., with respect to the random seed after running experiments multiple times)? \answerNo{}. We perform inference for diverse pre-trained models, which often do not come with pre-training code. Re-training all these models is not feasible. Additionally, LLM training is computationally expensive.
	\item Did you include the total amount of compute and the type of resources used (e.g., type of GPUs, internal cluster, or cloud provider)?
    \answerYes{} These details are provided in Section \ref{sec:experiments}
\end{enumerate}

\item If you are using existing assets (e.g., code, data, models) or curating/releasing new assets...
\begin{enumerate}
  \item If your work uses existing assets, did you cite the creators?
    \answerYes{}. We cite papers of all 7 datasets used in this work and also related models. Please see Section \ref{label:datasets}.
  \item Did you mention the license of the assets?
    \answerYes{}. See Appendix \ref{appendix:metadata} for details and Table \ref{tab:dataset_char} in Appendix.
  \item Did you include any new assets either in the supplemental material or as a URL?
    \answerYes{}
  \item Did you discuss whether and how consent was obtained from people whose data you're using/curating?
    \answerNA{}. We use publicly available datasets and strictly follow their provided license. 
  \item Did you discuss whether the data you are using/curating contains personally identifiable information or offensive content?
    \answerNA{}{}. We primarily use publicly available datasets. Most datasets are from less sensitive sources like Wikipedia and public financial transcripts, where we don't expect personally identifiable information or offensive content due to strict guidelines at source \url{https://en.wikipedia.org/wiki/Wikipedia:Offensive_material}.
\end{enumerate}

\item If you used crowdsourcing or conducted research with human subjects...
\begin{enumerate}
  \item Did you include the full text of instructions given to participants and screenshots, if applicable?
    \answerNA{}
  \item Did you describe any potential participant risks, with links to Institutional Review Board (IRB) approvals, if applicable?
    \answerNA{}
  \item Did you include the estimated hourly wage paid to participants and the total amount spent on participant compensation?
   \answerNA{}
\end{enumerate}

\end{enumerate}

%%%%%%%%%%%%%%%%%%%%%%%%%%%%%%%%%%%%%%%%%%%%%%%%%%%%%%%%%%%%
\appendix
We provide an overview of datasets employed, Limitation of the \dexter{} benchmark (Section \ref{appendix:limitations}), prompts (Section \ref{appendix:prompts}) and additional results in this Appendix.
\section{Data Card}
\textbf{Summary.} Section 3.1 and Table 1 in the main paper provide an overview of datasets in \dexter{} benchmark. We further provide example questions from each of the datasets in Table \ref{tab:datasets_overview:examples}. \\
\textbf{Languages.} English\newline
\textbf{Domain.} \dexter{} is an Complex QA benchmark with retrieval being evaluated in open-domain setup. the questions contained in the MusiqueQA (MQA), 2WikiMultihopqa (WQA), AmbigQA, OTT-QA and StrategyQA are from multiple domains. TAT-QA and FinQA questions focus on financial domain.

\textbf{Additional Details.} The benchmark is a curation of 7 datasets relevant to the domain of complex question answering. Table.\ref{tab:dataset_char} details each of these datasets with their homepage, licenses and characteristics. Section \ref{sec:dataset_structure} shows the structure of a sample with question, evidence and answer formed by our dataloader in \dexter{} toolkit. The Section also covers the json structure of raw data files. All the \textbf{data} can be found in form of zip files in the repository \textbf{\url{https://gitlab.tudelft.nl/venkteshviswan/bcqa_data}}. The corpus for all datasets except for AmbigQA, StrategyQA and OTT-QA are packaged along with dev, train and test questions in the zip files in the repo. The corpus files for AmbigQA, StartegyQA and OTT-QA are large and hence uploaded to a Google Drive and linked in readme of the repo \url{https://gitlab.tudelft.nl/venkteshviswan/bcqa_data}. A copy of our datasets would also be available in huggingface \url{https://huggingface.co/DEXTER-CQA}. We plan to continuously maintain the two data sources to ensure continuous development of the benchmark and to encourage open source contributions.

\textbf{Reproducibility}: For reproducibility developers can download the data files, install \dexter{} from source and run the inference scripts in evaluation folder from \dexter{} repo \url{https://github.com/VenkteshV/DEXTER}. Alternatively, we also provide a python package (pip install dexter-cqa) and an example jupyter notebook \url{https://colab.research.google.com/drive/1UOZ_JuDcWGKvwcPs4ygCEoGCUUgC1PUs?usp=sharing} to enable ease of access and reproducibility of results

\begin{table}[]
\resizebox{\textwidth}{!}{%
\begin{tabular}{|l|l|l|}
\hline
\textbf{Dataset Name} & \textbf{Homepage}& \textbf{Liscense}                                                  \\ \hline
MusiqueQA             & \url{https://github.com/StonyBrookNLP/musique}& \href{https://creativecommons.org/licenses/by/4.0/}{CC BY 4.0 License}                     \\ \hline
WikiMultiHopQA        & \url{https://github.com/Alab-NII/2wikimultihop} & \href{https://creativecommons.org/licenses/by/4.0/}{CC BY 4.0 License}                      \\ \hline
StrategyQA            & \url{https://allenai.org/data/strategyqa}& Provided under “MIT License” for non-commercial research purposes. \\ \hline
AmbigQA               & \url{https://nlp.cs.washington.edu/ambigqa/}& Creative Commons Attribution Share Alike 3.0                                                            \\ \hline
OTT-QA                & \url{https://ott-qa.github.io/}                 & Provided under “MIT License” for non-commercial research purposes. \\ \hline
TAT-QA                & \url{https://nextplusplus.github.io/TAT-QA/}    & \href{https://creativecommons.org/licenses/by/4.0/}{CC BY 4.0 License}                       \\ \hline
FinQA                 & \url{https://github.com/czyssrs/FinQA}& Provided under “MIT License” for non-commercial research purposes. \\ \hline
\end{tabular}%
}
\caption{ Characteristics of datasets included in DEXTER}
\label{tab:dataset_char}
\end{table}

\begin{table}[]
\resizebox{\textwidth}{!}{%
\begin{tabular}{|l|l|l|}
\hline
\textbf{Name} & \textbf{Component} & \textbf{Checkpoints}                                                                                                                                                                              \\ \hline
BM25          & Retrieval          & -                                                                                                                                                                                                 \\ \hline
SPLADE        & Retrieval          & \url{https://huggingface.co/naver/splade\_v2\_max}                                                                                                                         \\ \hline
DPR           & Retrieval          & \begin{tabular}[c]{@{}l@{}}Question: https://huggingface.co/facebook/dpr-question\_encoder-multiset-base\\ Context: https://huggingface.co/facebook/dpr-ctx\_encoder-multiset-base\end{tabular} \\ \hline
ANCE          & Retrieval          & \url{https://huggingface.co/sentence-transformers/msmarco-roberta-base-ance-firstp}                                                                                        \\ \hline
Tas-b         & Retrieval          & \url{https://huggingface.co/sentence-transformers/msmarco-distilbert-base-tas-b}                                                                                           \\ \hline
MPNet         & Retrieval          & \url{https://huggingface.co/sentence-transformers/multi-qa-mpnet-base-cos-v1}                                                                                              \\ \hline
Contriever    & Retrieval          & \url{https://huggingface.co/facebook/contriever}                                                                                                                           \\ \hline
ColBERTv2     & Retrieval          & \url{https://github.com/VenkteshV/ColBERT?tab=readme-ov-file\#overview}                                                                                                  \\ \hline
Flan-T5       & LLM                & \url{https://huggingface.co/google/flan-t5-xl}                                                                                                                             \\ \hline
Llama         & LLM                & \url{https://huggingface.co/meta-llama/Llama-2-7b-hf}                                                                                                                      \\ \hline
\end{tabular}%
}
\caption{Characteristics of models used in \dexter{}}
\label{tab:model_char}
\end{table}

\begin{table*}[ht!]
 \small
\begin{tabular}{lp{5cm}p{3cm}p{2cm}}
\hline
\textbf{Dataset} & \textbf{Example Question} & \textbf{Description} & \textbf{Ability} \\ \hline

\textbf{StrategyQA}~\cite{strategy_qa}  & \texttt{Does Andrew Johnson's presidential number exceed}      & comparison and  implicit questions &Reasoning over implicit aspects and commonsense reasoning\\
& \texttt{Elagabalus's Emperor number?}  &   &  \\
& \texttt{}  &   & \\

\textbf{WQA}~\cite{2wikimultihopqa} & \texttt{Who was born later, Gideon Johnson or Holm Jølsen?}      & comparison and compositional questions  & Compositional and Comparative Reasoning \\
 & & & \\

 \textbf{MQA}~\cite{trivedi-etal-2022-musique} &  \texttt{What did the actress in My Fair Lady win a Tony for ?}      &  compositional questions  &Compositional Reasoning\\

 \textbf{AmbigQA} \cite{min-etal-2020-ambigqa} &Who plays the doctor in dexter season 1? & Ambiguous QA& Detecting and reasoning under ambiguity \\
  & &  & \\

\colorg  \textbf{TAT-QA~\cite{tatqa}} & \colorg  What is the year on year percentage change in domestic discount rate between 2018 and 2019? & \colorg Table based numerical reasoning  &\colorg  Table Parsing,Text understanding, numerical reasoning  \\

\textbf{OTT-QA} \cite{ottqa} & What is the full birth name of the Bradford A.F.C player that only played for the team in 2011 ? & Table and Text based reasoning & Table Parsing, Text understanding \\

\textbf{FinQA}~\cite{chen2022finqa} & \texttt{In 2010 and 2009 , what was the total fair value in billions of assets segregated for the benefit of securities and futures brokerage customers? }      & Table and Text based numerical reasoning   & Table Parsing,Text understanding, numerical reasoning  \\

\bottomrule
\end{tabular}
\caption{Overview of the Complex QA  datasets used in this study. Abilities refers to skills required to solve the dataset.}
\label{tab:datasets_overview:examples}
\end{table*}

\subsection{Meta Information}
\label{appendix:metadata}
\textbf{Benchmark Curators.} The benchmark was created by Venktesh Viswanathan, Deepali Prabhu, and Avishek Anand.\newline
\textbf{Licensing Information.} The licensing information of each of the underlying datasets included in the benchmark is provided in Table.\ref{tab:dataset_char}. For dexter toolkit, we release the software with Apache 2.0 license.\newline
\textbf{Leader board/Benchmarks.} The benchmark is aimed to be used by researchers wanting to test various components of the RAG pipeline for the task of complex question answering. Currently, we have evaluated a range of retrievers and language models, as shown in the Results Section in the main paper. Our benchmark is also open to evaluation, and it is easy to add new retrieval and generative models. We have made the models and code used in our main experiments available on our GitHub repository. This allows others to reproduce our empirical results, develop their own models or datasets, extend our framework, and conduct meaningful evaluations. Users can easily access the available datasets by downloading the corresponding data files from our repository \url{https://gitlab.tudelft.nl/venkteshviswan/bcqa\_data} and utilizing the data loaders in DEXTER. Retrievers can be loaded using our retriever classes and the publicly available checkpoints. For LLM engines, users can either use their own API keys or the publicly available checkpoints. A summary of publicly used checkpoints for models is provided in Table.\ref{tab:model_char}. The users of DEXTER can easily extend any of its components to load custom datasets, retrievers or LLMs. We aim to ensure that the benchmark is not limited to our components or datasets alone, allowing for flexible and broad usage.

\subsection{Dataset Structure}
\label{sec:dataset_structure}
For benchmarking purposes, we have chosen to convert each of the base datasets into two types of files. The first file includes questions, their answers, context mappings, and metadata. The second file contains the  documents that contain the relevant contexts with their metadata. The projection to an open domain setting is further explained in Section 3.5 of the main paper. We use the train, test, and validation split configuration from the base dataset. The first file containing question-answer pairs has a free structure, with most of them following the original structure of the dataset. Listing.\ref{lst:data-example} shows a single sample formed by the data loaders in DEXTER. The file with the corpus has a fixed structure as shown in Listing.\ref{lst:corpus-example}. All the code to convert raw files from the base dataset to the standard format used in DEXTER is made on our GitHub repository.

\begin{lstlisting}[basicstyle=\ttfamily\footnotesize, frame=single, numbers=left, numberstyle=\tiny\color{gray}, keywordstyle=\color{blue}, caption={Example block showing standard structure used to form the corpus file.}, label={lst:corpus-example}]
[
  {
    "id": "idx",
    "title": "title of the context item",
    "text": "Main text of the context items",
    "type": "Type of the context item, currently table or text"
  },
  ...
]
\end{lstlisting}

\begin{lstlisting}[basicstyle=\ttfamily\footnotesize, frame=single, numbers=left, numberstyle=\tiny\color{gray}, keywordstyle=\color{blue}, caption={Example block showing standard structure of data formed by our data loaders}, label={lst:data-example}]
{
  "idx": "V/2008/page_17.pdf-1-0",
  "question": {
    "idx": "V/2008/page_17.pdf-1",
    "text": "what is the average payment volume per transaction
              for american express?"
  },
  "answer": {
    "idx": "V/2008/page_17.pdf-1",
    "text": "127.40"
  },
  "evidence": {
    "table": [
      [
        "visa inc. ( 1 )",
        "$ 2457",
        "$ 3822",
        "50.3",
        "1592"
      ],
      [
        "mastercard",
        "1697",
        "2276",
        "27.0",
        "916"
      ]
    ],
    "columns": [
      "company",
      "payments volume ( billions )"
    ],
    "idx": "V/2008/page_17.pdf-1.table"
  }
}
\end{lstlisting}

\section{Limitations}
\label{appendix:limitations}

While \dexter{} covers diverse aspects of complex QA tasks, it still has some limitations. Here we identify some limitations to aid in further enhancements to the benchmark and toolkit in the future. The enhancements would be continually handled and maintained by the core development team and would also be open to contributions.

\textbf{Multilingual Datasets}
While currently \dexter{} covers divers aspects of complexity and is a multi-domain benchmark, it is primarily composed of questions and documents in English. This is primarily due to lack of diverse multi-lingual complex QA datasets. In future, we plan to curate complex questions from diverse languages starting with integration of Mintaka \cite{sen-etal-2022-mintaka} dataset in \dexter{}.

\textbf{Evaluation on more open models}:
Currently \dexter{} supports OpenAI models and open-source models like Mistral, Llama2-7b and FlanT5. In our results, we have demonstrated the importance of scale of LLMs employed for reasoning. While we carry out the main experiments with gpt-3.5-turbo due to its superior performance, we also plan to support more open source models in the future. While it is difficult to catch up with rapid release of open models every week, we plan to add new models with considerable improvements on diverse benchmarks. We also plan to open our project for contributions to encourage evaluation of new models and methods.

\textbf{Toolkit enhancements and training support}: Our initial goals for \dexter{} was to comprehensively evaluate off-the-shelf retrieval models on complex QA tasks in open-domain setup. This was followed by evaluation of the impact of retrieval on LLM reasoning in a Retrieval Augmented Generation setup (RAG). Hence, the current toolkit does not support training of custom retrieval models. In future, we plan to incorporate this, in addition to exploring transformer based architectures that can extend to long context.
\section{Further Experimental Details}
\label{appendix:experimental}
\subsection{Retriever}
We evaluate diverse retrieval models with publicly available checkpoints, shown in Table \ref{tab:model_char}. Due to the length limit of transformer based models, we restrict the document length to first 512 word pieces, as done in prior work \cite{beir}. We select diverse retrieval models based on their unique characteristics. BM25 and SPLADE serve as strong lexical and sparse retrievers respectively. DPR is a well known off-the-shelf retrieval model employed for opne-domain QA. Tas-b is chosen due to the strong training objective that uses dual supervision. ANCE employs better sampling of negatives during training and MPNET is chosen due to different pre-training objective compared to BERT based language models. ColBERTv2 is a well known late-interaction model that mitigates the cost of cross-encoder by employing late interaction based attention mechanism while providing superior performance to bi-encoder based dense retrieval models. 

\subsection{Generative models}

We employ temperature of 0.3 to reduce randomness in generated outputs. We employ 5 few-shot samples to account for context length limitations in LLMs and to accommodate retrieved documents in RAG setup. The frequency and brevity penalty are set to 0.8 and 0.6 respectively. For all datasets except FinQA we evaluate on dev set due to lack of publicly available test sets with gold labels. For FinQA we evaluate on test set. While the original MusiqueQA dataset contains a large number of question in validations et filtering out unanswerable questions yields a validation set of 1252 questions \cite{self_ask}. For 2WikiMultiHopQA we perform retrieval and LLM inference for the first 1200 2-hop questions to mitigate cost of LLM inference as done in prior work \cite{trivedi-etal-2023-interleaving,self_ask}.
\begin{table*}[hbt!]
    \centering
    
    \begin{tabular}{lccccccccccccccc}
    \toprule
     \textbf{Method}&\multicolumn{1}{c}{WQA} & \multicolumn{1}{c}{MQA} & \multicolumn{1}{c}{AmbigQA} & \multicolumn{1}{c}{StrategyQA}& \multicolumn{1}{c}{Tat-QA} &\multicolumn{1}{c}{ FinQA} &\multicolumn{1}{c}{ OTT-QA} \\
   % &\multicolumn{1}{c|}{n@10} & \multicolumn{1}{c}{RR}&\multicolumn{1}{c|}{n@10} & \multicolumn{1}{c}{RR}&\multicolumn{1}{c|}{n@10} & \multicolumn{1}{c}{RR}&\multicolumn{1}{c|}{n@10} & \multicolumn{1}{c}{RR} &\multicolumn{1}{c|}{n@10} & \multicolumn{1}{c}{RR} &\multicolumn{1}{c|}{n@10} & \multicolumn{1}{c}{RR} &\multicolumn{1}{c|}{n@10} & \multicolumn{1}{c}{RR} \\
     % & cover-EM & cover-EM& EM& EM&EM \\
     \midrule

    \midrule
    
    \textbf{Lexical} & & \\
            BM25 &\textbf{0.416} & \textbf{0.259}& \textbf{0.637}&0.337&\textbf{0.794} & \textbf{0.429} & \underline{0.123} \\

     \midrule
     
      \textbf{Sparse} & & \\
            SPLADE & 0.323&0.181 & 0.620 & 0.272& \underline{0.749}& 0.361& 0.106 \\
     \midrule
      \textbf{Dense} & & \\
            DPR & 0.179& 0.107&0.560&0.124  &0.508&0.176&0.066      \\
            ANCE & 0.223&0.115 & 0.594 & 0.247&0.666&0.294&0.055 \\

                  %      MDR & & \\
                        tas-b & 0.303& 0.172&0.620&0.342&0.712&0.326&0.101\\
                        MPNet & 0.253 & 0.149 & 0.603& \textbf{0.355} & 0.683 & 0.326&0.129\\
       Contriever & 0.294 & 0.169&0.576&0.254&0.328&0.229&0.072 \\
    COlBERTV2 & \underline{0.333}  &\underline{0.211}&\underline{0.621}&\underline{0.339}&0.729& \underline{0.411}& \textbf{0.124}\\
\midrule
    % \textbf{Re-Ranking} & & & & & & & \\
    % BM25 + CE & & & & & & & \\
    % \bottomrule
%          \textbf{Semi-Oracle} & & & & & & & & & & \\
%         ClaimOnly & 33.03& 39.57& 36.31 & 58.15 & 33.81& 48.61& 25.70 & 23.99&28.55 & 63.79 & 7.95 & 33.42 &  43.70  \\

% \programfc{} & 38.57& 42.49 & 37.12  & 50.66& 35.22& 45.76& 33.43&32.50 & 32.95 & 55.11 & 25.44 & 37.83 & 43.79 \\
%     \claimdecomp{}&33.43 & 39.78 & 35.04&  55.49 & 33.93 & 48.53& 34.37& 33.50&29.48 & 63.11& 10.85 & 34.48& 44.19  \\
%      \numdecomp{}& 33.81& 39.46 &33.57 & 53.45& 34.18& 47.00& 35.23&34.23 &29.11 & 60.29 & 13.90 &34.43& 43.24\\
     \bottomrule
    \end{tabular}
    \caption{Dense retrieval results Recall@100 across complex QA benchmarks. The best results are highlighted in bold and second best results are underlined}
    % \vspace{-1cm}
    \label{tab:recall_100}
\end{table*}
\begin{table*}[hbt!]

    \small
 \begin{tabular}{lccccccccccc}
    \toprule
     \textbf{Method}& \multicolumn{1}{c}{MQA}& \multicolumn{1}{c}{WQA} & \multicolumn{1}{c}{AmbigQA} &\multicolumn{1}{c}{StrategyQA} &\multicolumn{1}{c}{Tat-QA} &\multicolumn{1}{c}{FinQA} & \multicolumn{1}{c}{OTT-QA}\\

      & c-EM & c-EM& $F1_{Ans}$&c-EM&c-EM & EM-Tol & c-EM \\
     \midrule
 
    \textbf{RAG-Oracle } & & & & & & & \\ 
    (\COT{}) & \\
        \ \ - gpt-3.5-turbo & \colorit \textbf{44.28} & \colorit \bf 65.55& 35.57  &  \colorit \bf73.36& 54.05&52.22 & \colorit \bf 46.23 \\ 
                  \ \ - Mistral-7b & \underline{22.54}& 25.00 &\underline{09.93} & \underline{62.55}&\underline{42.99} &\underline{12.55} &  \underline{18.39} \\ 
            
          \ \ - Llama2-7b  & 22.12 &\underline{47.33} & 9.09 &56.77  &30.46 &06.62 & 15.89\\

\midrule
    supervised baseline &  37.60  & 50.59 &  \colorit \textbf{42.30}  & 65.2   & \colorit \bf55.2 & \bf \colorit  58.86 & 33.20\\
     
     \bottomrule
    \end{tabular}
    \caption{QA Performance on \dexter{}. The model used  is gpt-3.5-turbo, unless otherwise specified. The best results are highlighted in bold. And second best results are underlined.}
    % \vspace{-1cm}
    %$\dagger$ and $\ddagger$ indicates statistical significance over \COT{} at 0.1 level  and 0.01 level respectively.
    \label{tab:llm_reasoning}
   % \vspace{-2em}
\end{table*}

 \begin{table*}[hbt!]

\begin{tcolorbox}[title= MusiqueQA Prompt]
\small
\textbf{Instruction}:\texttt{Follow the given examples and Given the question and context output final answer for the question using information in the context and give answer in form of  [Final Answer]: .}

\paragraph{\textbf{Exemplars}}:

% [Question]: \textit{what was the growth rate of reit index as of 12/31/2003}

% [Table]: ventas | $ 100 | $ 206 | $ 270 | $ 331 | $ 457 | $ 512 | 
% nyse composite index | $ 100 | $ 132 | $ 151 | $ 166 | $ 200 | $ 217 | 
% all reit index | $ 100 | $ 138 | $ 181 | $ 196 | $ 262 | $ 215 | 
% healthcare reit index | $ 100 | $ 154 | $ 186 | $ 189 | $ 273 | $ 279 | 
% russell 1000 index | $ 100 | $ 130 | $ 145 | $ 154 | $ 178 | $ 188 | 

% [Text]: \textit{stock performance graph the following performance graph compares the cumulative total return ( including dividends ) to the holders of our common stock from december 31 , 2002 through december 31 , 2007 , with the cumulative total returns of the nyse composite index , the ftse nareit composite reit index ( the 201call reit index 201d ) , the  \dots}

% [Program]: \textsf{ans=(( 138 - 100 )/100)*100}

% [Answer]: \textcolor{teal}{\textsf{38}}
% \\
% \dots \\
% \dots 
% \paragraph{\textbf{Test Input}}: Read the following table, and then answer the question:
% Table: {}
% Text: {}
% Question: {}

% Equation: [INS]
% Answer: [INS]

[Question]: What is the current official currency in the country where Ineabelle Diaz is a citizen? \\

[Answer]: Ineabelle Diaz is from Peurto Rico, which is in the United States of America. The current official currency in the United
States is the United States dollar. \\

[Final Answer]: United States dollar. \\

[Question]: Where was the person who founded the American Institute of Public Opinion in 1935 born? \\

[Answer]: The person who founded the American Institute of Public Opinion in 1935 is George Gallup. George Gallup was born 
in Jefferson, Iowa.  \\

[Final Answer]: Jefferson. \\

[Question]: What language is used by the director of Tiffany Memorandum? \\

[Answer]: The director of Tiffany Memorandum is Sergio Grieco. Sergio Grieco speaks Italian. \\

[Final Answer]: Italian. \\

[Question]: What is the sports team the person played for who scored the first touchdown in Superbowl 1? \\

[Answer]: The player that scored the first touchdown in Superbowl 1 is Max McGee. Max McGee played for the Green Bay
Packers. \\
                      
[\dots]  
\paragraph{\textbf{Test Input}}: Read the following context, and then answer the question:
Context: \{\}
Question: \{\}

Explanation: [INS]
Answer: [INS]

\end{tcolorbox}
\captionof{figure}{Example of In-context learning for MusiqueQA through manual \COT{} based prompting of LLMs (limited examples shown)}
\label{prompt:mqa} 
\end{table*}

 \begin{table*}[hbt!]

\begin{tcolorbox}[title= 2WikiMultiHopQA Prompt]
\small
\textbf{Instruction}:\texttt{Follow the given examples and Given the question and context output final answer for the question using information in the context and give answer in form of  [Final Answer]: .}

\paragraph{\textbf{Exemplars}}:

% [Question]: \textit{what was the growth rate of reit index as of 12/31/2003}

% [Table]: ventas | $ 100 | $ 206 | $ 270 | $ 331 | $ 457 | $ 512 | 
% nyse composite index | $ 100 | $ 132 | $ 151 | $ 166 | $ 200 | $ 217 | 
% all reit index | $ 100 | $ 138 | $ 181 | $ 196 | $ 262 | $ 215 | 
% healthcare reit index | $ 100 | $ 154 | $ 186 | $ 189 | $ 273 | $ 279 | 
% russell 1000 index | $ 100 | $ 130 | $ 145 | $ 154 | $ 178 | $ 188 | 

% [Text]: \textit{stock performance graph the following performance graph compares the cumulative total return ( including dividends ) to the holders of our common stock from december 31 , 2002 through december 31 , 2007 , with the cumulative total returns of the nyse composite index , the ftse nareit composite reit index ( the 201call reit index 201d ) , the  \dots}

% [Program]: \textsf{ans=(( 138 - 100 )/100)*100}

% [Answer]: \textcolor{teal}{\textsf{38}}
% \\
% \dots \\
% \dots 
% \paragraph{\textbf{Test Input}}: Read the following table, and then answer the question:
% Table: {}
% Text: {}
% Question: {}

% Equation: [INS]
% Answer: [INS]

[Question]: Who lived longer, Theodor Haecker or Harry Vaughan Watkins?? \\

[Answer]: ITheodor Haecker was 65 years old when he died. Harry Vaughan Watkins was 69 years old when he died. \\

[Final Answer]: Harry Vaughan Watkins. \\

[Question]: Why did the founder of Versus die? \\

[Answer]: The founder of Versus was Gianni Versace. Gianni Versace was shot and killed on the steps of his Miami Beach
mansion on July 15, 1997.  \\

[Final Answer]: Shot. \\

[Question]: Who is the grandchild of Dambar Shah? \\

[Answer]: TDambar Shah (? - 1645) was the king of the Gorkha Kingdom. He was the father of Krishna Shah. Krishna Shah
(? - 1661) was the king of the Gorkha Kingdom. He was the father of Rudra Shah. \\

[Final Answer]: Rudra Shah. \\

[Question]: Are both director of film FAQ: Frequently Asked Questions and director of film The Big Money from the same
country? \\

[Answer]:  The director of the film FAQ: Frequently Asked Questions is Carlos Atanes. The director of the film The Big
Money is John Paddy Carstairs. The nationality of Carlos Atanes is Spanish. The nationality of John Paddy Carstairs is
British \\
                      
[\dots]
\paragraph{\textbf{Test Input}}: Read the following context, think step by step and then answer the question:
Context: \{\}
Question: \{\}

Explanation: [INS]
Answer: [INS]

\end{tcolorbox}
\captionof{figure}{Example of In-context learning for 2WikiMultiHopQA through manual \COT{} based prompting of LLMs (limited examples shown)}
\label{prompt:wqa} 
\end{table*}

 \begin{table*}[]

\begin{tcolorbox}[title= FinQA Prompt]
\small
\textbf{Instruction}:\texttt{You are a helpful, respectful and honest assistant helping to solve math word problems or tasks requiring reasoning or math, using the information from given table and text.}

\paragraph{\textbf{Exemplars}}:

% [Question]: \textit{what was the growth rate of reit index as of 12/31/2003}

% [Table]: ventas | $ 100 | $ 206 | $ 270 | $ 331 | $ 457 | $ 512 | 
% nyse composite index | $ 100 | $ 132 | $ 151 | $ 166 | $ 200 | $ 217 | 
% all reit index | $ 100 | $ 138 | $ 181 | $ 196 | $ 262 | $ 215 | 
% healthcare reit index | $ 100 | $ 154 | $ 186 | $ 189 | $ 273 | $ 279 | 
% russell 1000 index | $ 100 | $ 130 | $ 145 | $ 154 | $ 178 | $ 188 | 

% [Text]: \textit{stock performance graph the following performance graph compares the cumulative total return ( including dividends ) to the holders of our common stock from december 31 , 2002 through december 31 , 2007 , with the cumulative total returns of the nyse composite index , the ftse nareit composite reit index ( the 201call reit index 201d ) , the  \dots}

% [Program]: \textsf{ans=(( 138 - 100 )/100)*100}

% [Answer]: \textcolor{teal}{\textsf{38}}
% \\
% \dots \\
% \dots 
% \paragraph{\textbf{Test Input}}: Read the following table, and then answer the question:
% Table: {}
% Text: {}
% Question: {}

% Equation: [INS]
% Answer: [INS]

 Text: annual sales of printing papers and graphic arts \\supplies and equipment totaled \$ 3.5 billion in 2012 compared with \$ 4.0 billion in 2011...
\\
\textbf{Table}: in millions | 2012 | 2011 | 2010 \\
sales | $ 6040 | $ 6630 | \$ 6735 \\
operating profit | 22 | 34 | 78 \\
\textbf{Question}: what percent of distribution sales where \\attributable to printing papers and graphic arts supplies and equipment in 2011?\\
\textbf{Rationale}: The sales of print papers and graphic arts \\ supplies and equipment in 2011 is 3.5 billion. The 
 \\ total sales in 2011 is 6.63 billion. The percentage is 52.8\%. So the answer is 52.8\%. \\
Answer:52.8%
\\ \\

\dots 
\paragraph{\textbf{Test Input}}: Read the following table,text and then answer the question:
Table: \{\}
Text: \{\}
Question: {}

Explanation: [INS]
Answer: [INS]

\end{tcolorbox}
\captionof{figure}{Prompt for FinQA}
\label{prompt:finqa} 
\end{table*}

\begin{table*}[hbt!]

\begin{tcolorbox}[title= StrategyQA Self-Ask Prompt]
\small
\textbf{Instruction}:\texttt{You are a helpful, respectful and honest assistant helping to solve commonsense problems requiring reasoning. Follow the given examples that use the facts to answer a question by decomposing into sub-questions first and then predicting the final answer as "Yes" or "No" only.}

\paragraph{\textbf{Exemplars}}:

[Facts]: Snowden scored above $145$ on two separate IQ tests. The minimum accepted IQ score for MENSA on the Stanford–Binet is $132$, while for the Cattell it is $148$.

[Question]: \textit{Could Edward Snowden join MENSA?}

[Sub-question 1]: \textsf{What is the minimum accepted IQ score to be admitted to MENSA?\\}
[Sub-question 2]: \textsf{What is Edward Snowden's IQ?\\}
[Sub-question 3]: \textsf{Is \#2 greater than or equal to \#1?}

[Answer]: \textcolor{teal}{\textsf{Yes}}
\\
\dots \\
\dots 
\paragraph{\textbf{Test Input}}: Facts: {} Question: {}

Sub-question: [INS]
Answer: [INS]

\end{tcolorbox}
\captionof{figure}{Prompt for StrategyQA}
\label{prompt:strategy} 
\end{table*}

\section{Prompts}
\label{appendix:prompts}

We show some of the manual \COT{} prompts used for RAG experiments for few datasets (Figures \ref{prompt:finqa},\ref{prompt:mqa},\ref{prompt:wqa}) and also an example of self-ask style prompting (Figure \ref{prompt:strategy}).  The few-shot samples used are applicable for closed-book, RAG top-k and RAG Oracle setups. The prompts comprise demonstration samples with rationales to elicit chain of thought reasoning in LLMs. All prompts employed for different OpenAI and open models on different datasets can be found in \dexter{} github repo \url{https://github.com/VenkteshV/DEXTER/tree/main/evaluation} in evaluation folder. The inference scripts for each datasets are housed under the ``llm" sub-folder. For instance, for RAG \COT{} inference using gpt-3.5-turbo the prompt can be found at \url{https://github.com/VenkteshV/DEXTER/blob/main/evaluation/musique/llms/run_gpt3_rag_few_SHOT.py}.

\section{Retrieval Results (Recall@100)}
\label{appendix:recall}
We report the Recall@100 for retrieval in Table \ref{tab:recall_100}. We observe that BM25 achieves high recall in 5/7 datasets. We also observe that ColBERTv2 provides the second-best recall results and is not far behind BM25 on most datasets. This demonstrates that pre-trained late-interaction based methods are more suited compared to other dense retrieval methods for complex QA tasks when used in an off-the-shelf manner. We also observe that sparse retrieval models like SPLADEv2 are also competitive and serve as a strong baseline. 

\section{Generative Model Results (Open Models)}
\label{appendix:open_llm}

We show the results for Llama2-7b in Table \ref{tab:llm_reasoning}. We observe that scale plays a major role when employing Large Language models for reasoning, as gpt-3.5-turbo vastly outperforms smaller models like Mistral and Lalam2 even when all models are provided with gold evidence. This demonstrates that scale of the LLM plays a major role in pronounced emergent capabilities like Chain of though and In-Context Learning. However, we observe immense potential for open source models to close this gap with stronger pre-training objectives, as their current performance on certain tasks like StrategyQA and TAT-QA are impressive . 

Among open source models we observe that mistral offers superior performance on 6/7 datasets in \dexter{} benchmark, outperforming Llama2 by upto \textbf{12 points} in TAT-QA. However, we observe that Llama2 has superior comparative reasoning capabilities, as it outperforms Mistral by 22 points. 

We observe that open source smaller models fall behind LLMs like gpt-3.5-turbo even in presence of gold evidence. Hence, our main experiments primarily employ gpt-3.5-turbo for diverse experimental setups. However, we observe there is immense potential for open source small LLMs in complex QA and hope \dexter{} would help contribute to further progress in this direction.

\end{document}